\pdfoutput=1

\documentclass[11pt]{article}

\usepackage[]{EACL2023}

\usepackage{times}
\usepackage{latexsym}

\usepackage[T1]{fontenc}

\usepackage[utf8]{inputenc}

\usepackage{microtype}

\usepackage{inconsolata}

\usepackage{xcolor}
\PassOptionsToPackage{dvipsnames}{xcolor}
\usepackage{multirow}
\usepackage{multicol}
\usepackage{graphicx}
\usepackage{tabu}
\usepackage{amsmath}
\usepackage{booktabs}
\usepackage{hyperref}

\usepackage{subcaption}
\usepackage{caption}
\usepackage[normalem]{ulem}
\usepackage{soul}
\usepackage[shortlabels]{enumitem}
\usepackage{array}
\usepackage{pgffor}
\usepackage{multirow}
\usepackage[normalem]{ulem}
\newcolumntype{P}[1]{>{\centering\arraybackslash}p{#1}}

\definecolor{light_purple}{HTML}{dfccfa}
\usepackage[most]{tcolorbox}
\tcbset{on line, 
        boxsep=1pt, left=0pt,right=0pt,top=0pt,bottom=0pt,
        colframe=white,
        colback=light_purple,  
        highlight math style={enhanced}
        }

\newcommand{\squad}{\textsc{SQuAD 1.1}}
\newcommand{\drop}{\textsc{DROP}}
\newcommand{\quoref}{\textsc{Quoref}}
\newcommand{\winogrande}{\textsc{Winogrande}}
\newcommand{\multirc}{\textsc{MultiRC}}
\newcommand{\ropes}{\textsc{Ropes}}

\newcommand{\sciqa}{\textsc{SciQA}}
\newcommand{\mctaco}{\textsc{MC-Taco}}
\newcommand{\piqa}{\textsc{PIQA}}
\newcommand{\qasc}{\textsc{QASC}}
\newcommand{\clariq}{\textsc{Clariq}}
\newcommand{\duorc}{\textsc{DuoRC}}
\newcommand{\hybridqa}{\textsc{HybridQA}}
\newcommand{\hotpotqa}{\textsc{HotpotQA}}
\newcommand{\cosmosqa}{\textsc{CosmosQA}}
\newcommand{\instructionbias}{\textit{instruction bias}}

\newcommand{\trn}{$\mathcal{S}_\text{train}$}
\newcommand{\trnp}{$\mathcal{S}_\text{train}^{p}$}
\newcommand{\trnnp}{$\mathcal{S}_\text{train}^{-p}$}
\newcommand{\tst}{$\mathcal{S}_\text{test}$}
\newcommand{\tstp}{$\mathcal{S}_\text{test}^{p}$}
\newcommand{\tstnp}{$\mathcal{S}_\text{test}^{-p}$}
\newcommand*\samethanks[1][\value{footnote}]{\footnotemark[#1]}

\newcommand{\nl}[1]{\textit{``#1''}}

\definecolor{green}{rgb}{0.4,0.7,0.0}
\definecolor{green_res}{rgb}{0.5,1.0,0.0}
\definecolor{red}{rgb}{1.0,0.2,0.0}

\title{Don't Blame the Annotator:\\Bias Already Starts in the Annotation Instructions}

\author{
Mihir Parmar$^{1}$\thanks{~~Equal Contribution} $\,$ \quad Swaroop Mishra$^{1}$\samethanks $\,$    \quad     \textbf{Mor Geva}$^{2}$\thanks{~~Now at Google Research} $\;$ \quad  \textbf{Chitta Baral}$^{1}$ $\;$ 
\\
 $^1$Arizona State University \;  $^2$Allen Institute for AI
 \\
 \small{\texttt{\{mparmar3, srmishr1, chitta\}@asu.edu, pipek@google.com}}
 }

\begin{document}
\maketitle

\begin{abstract}
In recent years, progress in NLU has been driven by benchmarks. These benchmarks are typically collected by crowdsourcing, where annotators write examples based on annotation instructions crafted by dataset creators.
In this work, we hypothesize that annotators pick up on patterns in the crowdsourcing instructions, which bias them to write many similar examples that are then over-represented in the collected data. We study this form of bias, termed \instructionbias{}, in 14 recent NLU benchmarks, showing that instruction examples often exhibit concrete patterns, which are propagated by crowdworkers to the collected data.
This extends previous work \cite{geva-etal-2019-modeling} and raises a new concern of whether we are modeling the \textit{dataset creator's instructions}, rather than the task.
Through a series of experiments, we show that, indeed, instruction bias can lead to overestimation of model performance, and that models struggle to generalize beyond biases originating in the crowdsourcing instructions. 
We further analyze the influence of instruction bias in terms of pattern frequency and model size, and derive concrete recommendations for creating future NLU benchmarks.\footnote{Code and data is available at \url{https://github.com/Mihir3009/instruction-bias}.}

\end{abstract}

\section{Introduction}

Benchmarks have been proven pivotal for driving progress in Natural Language Understanding (NLU) in recent years \cite{rogers2021qa, bach2022promptsource, wang2022benchmarking}. Nowadays, NLU benchmarks are mostly created through crowdsourcing, where crowdworkers write examples 
following annotation instructions crafted by dataset creators \cite{callison-burch-dredze-2010-creating, zheng2018crowdsourcing, suhr-etal-2021-crowdsourcing}. The instructions typically include a short description of the task, along with several examples \cite{dasigi2019quoref, zhou-etal-2019-going, sakaguchi2020winogrande}. 

Despite the vast success of this method, past studies have shown that data collected through crowdsourcing often exhibit various biases that lead to overestimation of model performance \cite{schwartz2017effect, gururangan-etal-2018-annotation, poliak2018hypothesis, tsuchiya-2018-performance, le2020adversarial, Mishra2020OurEM, mishra2021robust, hettiachchi2021investigating}. Such biases are often attributed to annotator-related biases, such as writing style and background knowledge \cite{gururangan-etal-2018-annotation, geva-etal-2019-modeling} (see more discussion on related work in \S\ref{sec:related_work}).

In this work, we propose that biases in crowdsourced NLU benchmarks often originate at an early stage in the data collection process of designing the annotation task. In particular, we hypothesize that task instructions 
provided by dataset creators, which serve as the guiding principles for annotators to complete the task, often influence crowdworkers to follow specific patterns, which are then propagated to the dataset and subsequently over-represented in the collected data. For instance, $\sim36\%$ of the instruction examples for the \quoref{} dataset \cite{dasigi2019quoref} start with \textit{``What is the name''}, and this same pattern can be observed in $\sim59\%$ of the collected instances.

To test our hypothesis, we conduct a broad study of this form of bias, termed \instructionbias{}, in 14 recent NLU benchmarks. We find that instruction bias is evident in most of these datasets, showing that $\sim73\%$ of instruction examples on average share a few clear patterns.
Moreover, we find that these patterns are propagated by annotators to the collected data, covering $\sim61\%$ of the instances on average. This suggests that instruction examples play a critical role in the data collection process and the resulting example distribution.

It is difficult to represent a task with a few examples, and bias in instruction examples makes it even more difficult since a task and its associated reasoning have a larger scope than instruction patterns. For example co-reference resolution, temporal commonsense reasoning, and numerical reasoning are much broader tasks than the prevalent patterns in \quoref{} (\textit{``what is the name...''}), \mctaco{} (\textit{``how long...''}) and \drop{} (\textit{``how many field goals...''}) datasets.

We investigate the effect of instruction bias on model performance, showing that performance is overestimated by instruction bias and that models often fail to generalize beyond instruction patterns. Moreover, we observe that a higher frequency of instruction patterns in the training set often increases the model performance gap on pattern and non-pattern examples and that large models are generally less sensitive to instruction bias.

In conclusion, our work shows that instruction bias widely exists in NLU benchmarks, often leading to an overestimation of model performance.
Based on our study, we derive concrete recommendations for monitoring and alleviating this bias in future data collection efforts.
From a broader perspective, our findings also have implications on the recent learning-by-instructions paradigm \cite{efrat2020turking, mishra2021cross}, where crowdsourcing instructions are used in model training.

\section{Instruction Bias in NLU Benchmarks}
\label{sec:instruction_bias}

Instructions are the primary resource for educating crowdworkers on how to perform their task \cite{nangia-etal-2021-ingredients}. Bias in the instructions, dubbed \instructionbias{}, could lead crowdworkers to propagate specific patterns to the collected data.

Here, we study instruction bias in NLU benchmarks\footnote{All benchmarks are in English.}, focusing on two research questions: 
(a) Do crowdsourcing instructions exhibit patterns that annotators can pick up on? 
and (b) Are such patterns propagated by crowdworkers to the collected data? 
In our study, we use the instructions of 14 recent NLU benchmarks:\footnote{The instructions were obtained from \citet{mishra2021cross}, who have collected those from the dataset authors.} (1) \clariq{} \cite{aliannejadi2020convai3}, (2) \cosmosqa{} \cite{huang2019cosmos}, (3) \drop{} \cite{dua2019drop}, (4) \duorc{} \cite{saha2018duorc}, and (5) \hotpotqa{} \cite{yang2018hotpotqa}  (6) \hybridqa{} \cite{chen2020hybridqa}, (7) \mctaco{} \cite{zhou-etal-2019-going}, (8) \multirc{} \cite{khashabi2018looking}, (9) \piqa{} \cite{bisk2020piqa}, (10) \qasc{} \cite{khot2020qasc}, (11) \quoref{} \cite{dasigi2019quoref}, (12) \ropes{} \cite{lin2019reasoning}, (13) \sciqa{} \cite{welbl2017crowdsourcing}, (14) \winogrande{} \cite{sakaguchi2020winogrande}. 
These benchmarks were created through different crowdsourcing protocols to evaluate diverse tasks \cite{mishra2021cross} (see dataset statistics in \S\ref{app:dataset_stats}).

\begin{table}[t!]
    \setlength \tabcolsep{2.0pt}
    \setlength{\belowcaptionskip}{-10pt}
    \centering
    \footnotesize
    \begin{tabular}{p{1.6cm}p{2.5cm}ccc}
        \textbf{Dataset} & \textbf{Pattern} & \textbf{\% Ins.} &  \textbf{\% $\mathcal{S}_\text{train}$} &  \textbf{\% $\mathcal{S}_\text{test}$} \\
        \midrule
        \clariq & \texttt{[Are|Would|Do] you} & 72.2 & 85.1 & 89 \\
        \midrule
        \cosmosqa & \texttt{What AUX} & 87.5 & 45.1 & 38.4 \\
        \midrule
        \drop & \texttt{How many [field goals | years | yards | points | touchdowns]} & 70 & 62.5 & 62.5 \\
        \midrule
        \duorc & \texttt{[How old | How | What | Who] AUX} & 70 & 85.1 & 84 \\
        \midrule
        \hotpotqa & \texttt{[In | Of | From | \_ ] [Which|What] AUX} & 87.5 & 53.8 & 54.2 \\
        \midrule
        \hybridqa & \texttt{Which AUX} & 29.4 & 25.7 & 15.1 \\
        \midrule
        \multirow{5}{*}{\mctaco} 
        & \texttt{How long AUX} & 100 & - & 87.6 \\
        \cmidrule{2-5}
        & \texttt{What AUX} & 100 & - & 90.1 \\
        \cmidrule{2-5}
        & \texttt{How often AUX} & 100 & - & 85.3 \\
        \cmidrule{2-5}
        & \texttt{AUX... [still|always |by the time] } & 100 & - & 67.3 \\
        \cmidrule{2-5}
        & \texttt{When did / What time} & 100 & - & 83.4 \\
        \midrule
        \multirc & \texttt{What AUX} & 14.3 & 38.4 & 41.5 \\
        \midrule
        \piqa & \texttt{How [do|can]} & 66.7 & 43.7 & 42.9 \\
        \midrule
        \qasc & \texttt{What AUX} & 57.1 & 49.3 & 47 \\
        \midrule
        \quoref & \texttt{What AUX~the [\_| full|real|first |last] name} & 36.4 & 57 & 60  \\
        \midrule
        \ropes & \texttt{Which AUX} & 42.9 & 74.1 & 20.7 \\
        \midrule
        \sciqa & \texttt{What AUX} & 100 & 83.6 & 84.5 \\
        \midrule
        \textsc{Wino- grande} & \texttt{[because | so | while | since | but] ... the} & 73.7 & 63.4 & 63.1 \\
        \midrule
        \textbf{Average} & ~ & 72.7 & 59 & 62 \\
    \end{tabular}
    \caption{Portion of patterns in instruction examples (Ins.) and in the corresponding train ($\mathcal{S}_\text{train}$) and test ($\mathcal{S}_\text{test}$) sets of NLU datasets. $\texttt{AUX} \in \{$am, is, are, was, were, has, have, had, do, does, did, will, would, can, could, may, might, shall, should, must$\}$, and \texttt{\_} is an empty string. 
    \mctaco{} has 5 different data subsets corresponding to different types of temporal reasoning (see Tab.~\ref{tab:artifacts_stats}), hence, the sum of percentages for this dataset exceeds 100\%.}
    \label{tab:artifacts}
\end{table}

Tab.~\ref{tab:artifacts_stats} provides the number of examples present in crowdsourcing instructions of each dataset. From Tab.~\ref{tab:artifacts_stats}, we can observe that our analysis involves a wide range of different tasks. Also, we believe that the lower number of examples in crowdsourcing instructions might be limiting the imagination of annotators while creating samples, resulting in instruction bias.

\begin{table}[t!]
\centering
\resizebox{\linewidth}{!}{

\begin{tabular}{c c c}
\toprule
\textbf{Dataset} & \textbf{Task} & \textbf{\# of Examples} \\
\midrule
\clariq & Clarification QA & 18 \\
\midrule
\cosmosqa & Commonsense Reasoning & 8 \\
\midrule
\drop & Numerical Reasoning & 10 \\
\midrule
\duorc & Paraphrased RC & 10 \\ 
\midrule
\hotpotqa & Multi-hop QA & 8 \\
\midrule
\hybridqa & QA & 17 \\
\midrule
\multirow{5}{*}{\mctaco} 
& Event Duration & 3  \\
\cmidrule{2-3}
& Event Ordering & 2 \\
\cmidrule{2-3}
& Frequency & 2 \\
\cmidrule{2-3}
& Stationary & 2 \\
\cmidrule{2-3}
& Absolute Point & 2 \\
\midrule
\multirc & Complex QA & 7 \\
\midrule
\piqa & Physical Interaction QA & 6 \\
\midrule
\qasc & Complex QA & 7 \\
\midrule
\quoref & Coreference QA & 11 \\
\midrule
\ropes & RC & 14 \\
\midrule
\sciqa & Science-based QA & 6 \\
\midrule
\winogrande & Commonsense Reasoning & 19 \\
\midrule
\textbf{Average} & ~ & 8.4 \\
\end{tabular}
}
\caption{Tasks of each dataset and number of examples in crowdsourcing instruction of each dataset. RC: Reading Comprehension, QA: Question Answering.}
\label{tab:artifacts_stats}
\end{table}

\subsection{Patterns in Crowdsourcing Instructions} 
\label{subsec:instructions_bias}

Our goal is to quantify biases in instruction examples that propagate to collected data instances. In this study, we focus on an intuitive form of bias of recurring word patterns, which crowdworkers can easily pick up on. 
To find such patterns, we manually analyze the instruction examples of each dataset to find a \textit{dominant pattern}, using the following procedure: 
(a) identifying repeating patterns of $n \geq 2$ words, (b) merging patterns that are semantically similar or have a substantial word overlap, and (c) selecting the most frequent pattern as the dominant pattern (an example is provided in \S\ref{sec:extraction_method}).

Tab.~\ref{tab:artifacts} shows the dominant pattern in the instruction examples of each dataset. 
On average, $72.7\%$ of the instruction examples used to create a dataset exhibit the same dominant pattern, and for 10 out of 14 datasets, the dominant pattern covers more than half of the instruction examples.
This suggests that crowdsourcing instructions demonstrate a small set of repeating ``shallow'' patterns. Moreover, the short length of the patterns (2-4 words) and the typical low number of instruction examples (Tab.~\ref{tab:artifacts_stats}) make the patterns easily visible to crowdworkers, who can end up following them.


Notably, our results are an underestimation of the actual instruction bias, since (a) we only consider the dominant pattern for each dataset (b) our manual analysis over instruction examples has a preference for short patterns (c) we do not consider paraphrased patterns (beyond the shallow paraphrases which are visible in annotation instructions), and (d) datasets may include implicit patterns (e.g. writing style and biases from the annotator’s background knowledge) that also contribute to instruction bias. Accounting for such patterns is expected to increase the bias percentage in Tab.~\ref{tab:artifacts} further.

\subsection{Instruction Bias Propagation to Datasets} 
\label{subsec:bias_propagation}

We now turn to investigate whether patterns in instruction examples are further propagated by crowdworkers to the collected data.
We analyze the train and test sets of each benchmark\footnote{If no explicit test set exists, we use the validation set.} to find the same patterns, using simple string matching. To account for syntactic modifications in identified patterns based on some examples from dataset, we also consider synonym words where appropriate and match the paraphrased version of each pattern.

Tab.~\ref{tab:artifacts} shows the results. Across all datasets, instruction patterns are ubiquitous in the collected data, occurring in $60.5$\% of the instances on average, with similar presence in training (59\%) and test (62\%) examples. 
While the dominant pattern's frequency in the data is typically not higher than in the instructions, for \clariq{}, \duorc{}, \multirc{}, \quoref{} and \ropes{}, the pattern frequency was amplified by the crowdworkers. Interestingly, these datasets used a relatively large number of instruction examples (Tab.~\ref{tab:artifacts_stats}), suggesting that more examples do not necessarily alleviate the propagation of instruction bias.
Example data instances with instruction patterns are provided in \S\ref{app:artifacts_examples}.

A natural question that arises is whether patterns in collected data reflect the true task distribution rather than a bias in the instructions. We argue that this is highly unlikely. 
First, while the space of possible patterns for a NLU task is arguably large, the dataset patterns are imbalanced proportionately to the patterns in the instructions for collecting it. 
For example, questions for assessing temporal commonsense reasoning could have various forms, such as \nl{What is the duration of...}, \nl{For how much time...}, and \nl{How long...}. However, \mctaco{} (event duration) is heavily dominated (87.6\%) by questions with the pattern \nl{how long}, which appears in 100\% of the instruction examples (Tab.~\ref{tab:artifacts}). 
In addition, datasets of similar tasks have different dominant patterns, while each dataset's dominant pattern correlates with the pattern in the corresponding instructions; the pattern `What is' appears in 48\% of the questions in \qasc{} and in 57\% of the instruction examples, but it is entirely different from the dominant pattern of \hotpotqa{}, which is another QA dataset for multi-hop reasoning.

We further validate the propagation of bias in instruction examples
by comparing the pattern distributions of collected instances when the instructions include and do not include examples.
We conduct this experiment for \mctaco{} and \quoref{} and find that, 
without any examples provided, the dominant pattern is substantially less frequent, showing that instruction bias is propagated during data collection. Full details are provided in \S\ref{app:survey}.

Propagation of instruction bias to the test set raises concerns regarding its reliability for evaluation,
which we address next.

\begin{table}[t]
    \setlength\tabcolsep{4.0pt}
    \setlength{\belowcaptionskip}{-10pt}
    \centering
    \footnotesize
    \begin{tabular}{lcc|cc}
         & \multicolumn{2}{c}{\textbf{Base}} & \multicolumn{2}{c}{\textbf{Large}} \\
         \cmidrule{2-5}
         & \tstp{} & \tstnp{} & \tstp{} & \tstnp{} \\
         \toprule
         \clariq{} & 30.5 & 27.4 \tcbox{\footnotesize{10.2\% $\downarrow$}} & 30.3 & 26.3 \tcbox{\footnotesize{13.2\% $\downarrow$}} \\
         \drop{} & 75.7 & 31.8 ~~~\tcbox{\footnotesize{58\% $\downarrow$}} & & \\
         \multirc{} & 40.5 & 33.7 \tcbox{\footnotesize{16.8\% $\downarrow$}} & ~~42 & 37.4 ~~~\tcbox{\footnotesize{11\% $\downarrow$}} \\
         \piqa{} & 20.7 & ~~~15 \tcbox{\footnotesize{27.5\% $\downarrow$}} & 21.8 & 15.3 \tcbox{\footnotesize{29.8\% $\downarrow$}} \\
         \quoref{} & 85.9 & ~~~66 \tcbox{\footnotesize{23.2\% $\downarrow$}} & 92.1 & 81.1 \tcbox{\footnotesize{11.9\% $\downarrow$}} \\
         \ropes{} & ~~57 & ~~~42 \tcbox{\footnotesize{26.3\% $\downarrow$}} & 57.8 & 58.2 ~~\tcbox{\footnotesize{0.7\% $\uparrow$}} \\
         \sciqa{} & 80.7 & 79.7 ~~\tcbox{\footnotesize{1.2\% $\downarrow$}} & 82.8 & 81.9 ~~\tcbox{\footnotesize{1.1\% $\downarrow$}} \\
         \midrule
         \textbf{Average} & 55.9 & 42.2 \tcbox{\footnotesize{24.5\% $\downarrow$}} & 54.5 & ~~50 ~~~\tcbox{\footnotesize{8.3\% $\downarrow$}} \\
    \end{tabular}
    \caption{Performance on \tstp{} vs. \tstnp{} of models trained on data instances containing instruction patterns (\trnp{}).}
\label{tab:just_artifacts}
\end{table}

\section{Effect on Model Learning}
\label{sec:model_learning}


Let \trn{} (\tst{}) be the set of training (test) examples, and denote by \trnp{} (\tstp{}) and \trnnp{} (\tstnp{}) its disjoint subsets of examples with and without instruction patterns, respectively.
We conduct two experiments where we fine-tune models on (a) \trnp{} and (b) \trnp{}$\cup\,$\trnnp{}, and evaluate them on \tstnp{} and \tstp{}.
This is to assess to what extent models generalize from instruction patterns to the downstream task (a), and to compare model performance on instances with and without instruction patterns (b).

\subsection{Experimental Setting}
\label{subsec:experimental_setting}

\paragraph{Datasets} 
Since model training is computationally expensive, we select a subset of seven datasets from those analyzed in \S\ref{sec:instruction_bias}: (1) \clariq{}, (2) \drop{}, (3) \multirc{}, (4) \piqa{}, (5) \quoref{}, (6) \ropes{}, and (7)  \sciqa{}. 
These datasets cover a variety of tasks, different types and levels of instruction bias (Tab.~\ref{tab:artifacts}), and are different in size (\S\ref{app:dataset_stats}).

\paragraph{Models} 
For all datasets except \drop{}, we evaluate T5-base and T5-large \cite{raffel2019exploring}, and BART-base and BART-large \cite{lewis-etal-2020-bart}. For \drop{}, we use Numnet+ \cite{ran2019numnet}, a RoBERTa model \cite{liu2019roberta} with specialized output heads for numerical reasoning. Numnet+ has 355M parameters, which is closer to T5-base (220M) than to T5-large (770M) in size. 

\paragraph{Evaluation} We evaluate model performance using the standard F$_1$ evaluation score, and report the average score over three random seeds. 


\begin{table}[t]
    \setlength\tabcolsep{4.0pt}
    \setlength{\belowcaptionskip}{-10pt}
    \centering
    \footnotesize
    \begin{tabular}{lcc|cc}
         & \multicolumn{2}{c}{\textbf{Base}} & \multicolumn{2}{c}{\textbf{Large}} \\
         \cmidrule{2-5}
         & \tstp{} & \tstnp{} & \tstp{} & \tstnp{} \\
         \toprule
         \clariq{} & 30.5 & 26.9 \tcbox{\footnotesize{11.8\% $\downarrow$}} & 30.7 & 26.9 \tcbox{\footnotesize{12.4\% $\downarrow$}} \\
         \drop{} & 76 & 78.9 ~~\tcbox{\footnotesize{3.8\% $\uparrow$}} & & \\
         \multirc{} & 40.5 & ~ 39 ~~~\tcbox{\footnotesize{3.7\% $\downarrow$}} & 42.4 & 44.5 ~~~~~\tcbox{\footnotesize{5\% $\uparrow$}} \\
         \piqa{} & 20.7 & 19.8 ~~\tcbox{\footnotesize{4.4\% $\downarrow$}} & 21.9 & 20.6 ~~\tcbox{\footnotesize{5.9\% $\downarrow$}} \\
         \quoref{} & 86.7 & 73.1 \tcbox{\footnotesize{15.7\% $\downarrow$}} & 92.1 & 81.1 \tcbox{\footnotesize{11.9\% $\downarrow$}} \\
         \ropes{} & 59.9 & 48.7 \tcbox{\footnotesize{18.7\% $\downarrow$}} & 59 & 64.2 ~~\tcbox{\footnotesize{8.8\% $\uparrow$}} \\
         \sciqa{} & 80.6 & 80.2 ~~\tcbox{\footnotesize{0.5\% $\downarrow$}} & 82.8 & 82.9 ~~\tcbox{\footnotesize{0.1\% $\uparrow$}} \\
         \midrule
         \textbf{Average} & 56.4 & 52.4 ~~\tcbox{\footnotesize{7.1\% $\downarrow$}} & 54.8 & 53.4 ~~\tcbox{\footnotesize{2.6\% $\downarrow$}} \\
    \end{tabular}
    \caption{Performance on \tstp{} vs. \tstnp{} of models trained on \trn{}.}
\label{tab:full_resutls}
\end{table}

\subsection{Results}
\label{subsec:model_learning_results}

We observe similar results for T5 and BART, and thus, present only the results for T5 in this section. Results for BART are provided in \S\ref{app:additional_results}.

\paragraph{Models often fail to generalize beyond instruction patterns.}
Tab.~\ref{tab:just_artifacts} shows the performance on \tstp{} and \tstnp{} when training only on examples with instruction patterns.
Across all experiments, there are large performance gaps, reaching to $58\%$ in \drop{} and $>10\%$ in both base and large models for \clariq{}, \multirc{}, \piqa{}, and \quoref{}. This indicates that models trained only on examples with instruction patterns fail to generalize to other task examples, and stresses that instruction bias should be monitored and avoided during data collection.
Notably, the gap is lower for large models than for base ones, showing that large models are less sensitive to instruction bias. This might be attributed to their larger capacity to capture knowledge and skills during pre-training.

\paragraph{Model performance is overestimated by instruction bias.}
We compare the performance on \tstp{} and \tstnp{} of models trained on the full training set (Tab.~\ref{tab:full_resutls}).
The average performance across all datasets is higher on examples that exhibit instruction patterns by $\sim7\%$ and $\sim3\%$ for the base and large models, respectively. Specifically, base models perform worse on \tstnp{} than on \tstp{} for all datasets except \drop{}, in some cases by a dramatic gap of $>15\%$ (e.g. $18.7\%$ in \ropes{} and $15.7\%$ in \quoref{}). In contrast, results for the large models vary across datasets, while the performance gap is generally smaller in magnitude. This shows that model performance is often overestimated by instructions bias, and reiterates that large models are generally less sensitive to instruction patterns. 


\section{Conclusions and Discussion}

We identify a prominent source of bias in crowdsourced NLU datasets, called \instructionbias{}, which originates in annotation instructions written by dataset creators.
We study this bias in 14 NLU benchmarks, showing that instruction examples used to create NLU benchmarks often exhibit clear patterns that are propagated by annotators to the collected data.
In addition, we investigate the effect of instruction bias on model performance, showing that instruction patterns can lead to overestimated performance as well as limit the ability of models to generalize to other task examples.

Based on our findings, we derive three recommendations for future crowdsourced NLU benchmarks: (1) Crowdsourcing instructions should be diverse; this could be achieved, for example, by having a large number of instructive examples, rephrasing examples using neural models, or periodically sampling examples from a \textit{diverse} set of previously collected examples. The latter could be done, for example, by maintaining a pool of diverse examples during the collection process and then presenting every annotator with a different random sample from this growing pool. (2) Word patterns in collected instances should be analyzed during data collection, as well as possible correspondence to instruction examples. Such analysis will help researchers monitor the collection process and the quality of the resulting data. (3) Correlation between model performance and input patterns should be checked during evaluation. 

\section*{Limitations}

This work covers 14 NLU datasets, for which annotation instructions are publicly available. However, most of these datasets are QA datasets. Our analysis can be extended to other NLU task categories, such as Natural Language Inference (NLI) and Relation Extraction (RE).

Our study reveals a concrete bias that skews the collected data distribution toward specific patterns. While the effect of instruction examples on collected data is prominent, it is hard to quantify how different the distribution of crowdsourced examples is from the natural distribution of the task. Concretely, to conduct a study that compares the distributions of crowdsourced versus natural complex reasoning questions, datasets of complex natural questions are needed. However, to the best of our knowledge, as of today, no such datasets exist.

In our analysis, we focused on shallow patterns based on word matching, however, it is known that there are other types of biases that are implicit in the text. Exploring these kinds of biases can be an interesting future direction. In addition, our analysis of model performance is based on splitting dataset instances based on the dominant pattern. However, it might be possible that there are more patterns, and the non-pattern subset might include other less frequent patterns. Hence, exploring the effect of different less frequent patterns on model learning can be a future work. 

Last, our work studied the effect of instruction bias on widely used generative models (i.e., T5 and BART); it would be valuable to investigate whether our findings hold in encoder-only models, such as BERT \cite{devlin-etal-2019-bert} and RoBERTa \cite{liu2019roberta}.

\section*{Acknowledgement}
We thank Daniel Khashabi, Sewon Min, and Avia Efrat for helpful feedback, and the anonymous reviewers for constructive suggestions.
We acknowledge the Research Computing (RC) at Arizona State University (ASU) for providing computing resources for experiments.

\bibliography{anthology,custom}
\bibliographystyle{acl_natbib}

\clearpage

\appendix

\section{Biases in NLU Benchmarks}
\label{sec:related_work}

Crowdsourcing has been a widely adapted approach to create large scale datasets such as \squad \cite{rajpurkar-etal-2016-squad, rajpurkar-etal-2018-know}, \drop \cite{dua2019drop}, \quoref \cite{dasigi2019quoref} and many more \cite{najafabadi2015deep, callison-burch-dredze-2010-creating, lasecki2014glance, zheng2018crowdsourcing, chang2017revolt}. Many past works investigate different types of bias in crowdsourcing datasets such as cognitive bias \cite{eickhoff2018cognitive}, annotator bias \cite{gururangan-etal-2018-annotation, geva-etal-2019-modeling}, sampling bias \cite{hu2020crowdsourcing}, demographic bias \cite{rahmani2021demographic} and others \cite{hettiachchi2021investigating}. Many works on bias in NLU benchmarks focus on biases resulting from the crowdsourcing annotations, and how annotator-specific patterns create biases in data \cite{geva-etal-2019-modeling}. 

To mitigate the bias, prior works have focused on priming crowdsourcing annotators with minimal information to increase their imagination \cite{Geva2021DidAU, clark2020tydi} to avoid recurring patterns. \citet{arunkumar2020real} develops a real time feedback and metric-in-the loop \cite{mishra2020dqi} workflow to educate crowdworkers in controlling dataset biases. \citet{nangia-etal-2021-ingredients} provides an iterative protocol with expert assessments for crowdsourcing data collection to increase difficulty of instances. \cite{swayamdipta2020dataset} introduces dataset map as a model-based tool to characterize and diagnose datasets. Also, \citet{karimi-mahabadi-etal-2020-end, mahabadi2021variational} propose learning strategies to train neural models, which are more robust to such biases and transfer better to out-of-domain datasets.

In this work, we show that biases exhibited by annotators start from the crowdsourcing instructions designed by dataset creators.



\begin{table*}[ht]

\setlength\tabcolsep{4.0pt}
\centering
\footnotesize

\begin{tabular}{lccc|ccc}

\multirow{2}{*}{\textbf{Dataset}} & \multicolumn{3}{c}{\textbf{Train}}                                                                                                       & \multicolumn{3}{c}{\textbf{Test}}                                                                                                       \\ \cmidrule{2-7} 
                                  & \multicolumn{1}{c}{$\mathcal{S}_\text{train}$} & \multicolumn{1}{c}{\trnp{}}                                   & \trnnp{}                             & \multicolumn{1}{c}{$\mathcal{S}_\text{test}$} & \multicolumn{1}{c}{\tstp}                                  & \tstnp                               \\ \midrule
\clariq                            & 8566 & \multicolumn{1}{c}{~~7286 \tcbox{\footnotesize{85.1\%}}}  & ~1280 \tcbox{\footnotesize{14.9\%}} & 4499 & \multicolumn{1}{c}{~~~4006 \tcbox{\footnotesize{89\%}}} & ~~~~493 \tcbox{\footnotesize{11\%}}  \\
\drop                              & 77409 & \multicolumn{1}{c}{48422 \tcbox{\footnotesize{62.5\%}}}  & 28987 \tcbox{\footnotesize{37.5\%}} & 9536 & \multicolumn{1}{c}{~5960 \tcbox{\footnotesize{62.5\%}}}  & 3576 \tcbox{\footnotesize{37.3\%}}  \\
\multirc                           & 5131 & \multicolumn{1}{c}{~~1972 \tcbox{\footnotesize{38.4\%}}}  & ~3159 \tcbox{\footnotesize{61.6\%}} & 953 & \multicolumn{1}{c}{~~395 \tcbox{\footnotesize{41.5\%}}}  & ~~558 \tcbox{\footnotesize{58.6\%}}  \\
\piqa                              & 17171 & \multicolumn{1}{c}{~~7508 \tcbox{\footnotesize{43.7\%}}}  & ~9663 \tcbox{\footnotesize{56.3\%}} & 3268 & \multicolumn{1}{c}{1401 \tcbox{\footnotesize{42.9\%}}} & 1867 \tcbox{\footnotesize{57.1\%}} \\
\quoref                            & 19399 & \multicolumn{1}{c}{~~~11052 \tcbox{\footnotesize{57\%}}} & ~~8347 \tcbox{\footnotesize{43\%}} & 2418 & \multicolumn{1}{c}{~~~1451 \tcbox{\footnotesize{60\%}}} & ~~~~967 \tcbox{\footnotesize{40\%}}  \\
\ropes                             & 1412 & \multicolumn{1}{c}{~~1046 \tcbox{\footnotesize{74.1\%}}}  & ~~366 \tcbox{\footnotesize{25.9\%}}  & 203 & \multicolumn{1}{c}{~~~42 \tcbox{\footnotesize{20.7\%}}}   & ~161 \tcbox{\footnotesize{79.3\%}}  \\
\sciqa                             & 11679 & \multicolumn{1}{c}{9765 \tcbox{\footnotesize{83.61\%}}}  & ~1914 \tcbox{\footnotesize{16.4\%}} & 1000 & \multicolumn{1}{c}{~845 \tcbox{\footnotesize{84.5\%}}}  & ~155 \tcbox{\footnotesize{15.5\%}}  \\ \midrule
\textbf{Total}                    & 140767 & \multicolumn{1}{c}{87051 \tcbox{\footnotesize{61.8\%}}}  & 53716 \tcbox{\footnotesize{38.2\%}} & 21877 & \multicolumn{1}{c}{14100 \tcbox{\footnotesize{64.5\%}}}  & 7777 \tcbox{\footnotesize{35.6\%}}
\end{tabular}

\caption{Statistics of number of train and test examples with and without instruction patterns. $\mathcal{S}_\text{train}$: set of examples in train set, \trnp{}: set of examples in train set with instruction pattern, \trnnp{}: set of examples in train set without instruction pattern, $\mathcal{S}_\text{test}$: set of examples in test set, \tstp{}: set of examples in test set with instruction pattern, \tstnp{}: set of examples in test set without instruction pattern.}
\label{tab:dataset_stats}
\end{table*}

\begin{table}[t]
    \setlength\tabcolsep{4.0pt}
    \setlength{\belowcaptionskip}{-10pt}
    \centering
    \footnotesize
    \begin{tabular}{lcc|cc}
         & \multicolumn{2}{c}{\textbf{Base}} & \multicolumn{2}{c}{\textbf{Large}} \\
         \cmidrule{2-5}
         & \tstp{} & \tstnp{} & \tstp{} & \tstnp{} \\
         \toprule
         \clariq{} & ~~30 & 26.2 \tcbox{\footnotesize{12.7\% $\downarrow$}} & 29.2 & 25.6 \tcbox{\footnotesize{12.3\% $\downarrow$}} \\
         \multirc{} & 27.9 & 15.1 \tcbox{\footnotesize{45.9\% $\downarrow$}} & 31.1 & ~~~21 \tcbox{\footnotesize{32.5\% $\downarrow$}} \\
         \piqa{} & 21.9 & 15.3 \tcbox{\footnotesize{30.1\% $\downarrow$}} & ~~23 & 16.1 ~~~\tcbox{\footnotesize{30\% $\downarrow$}} \\
         \quoref{} & 78.8 & 45.8 \tcbox{\footnotesize{41.9\% $\downarrow$}} & ~~87 & 58.6 \tcbox{\footnotesize{32.6\% $\downarrow$}} \\
         \ropes{} & 43.9 & 33.1 \tcbox{\footnotesize{24.6\% $\downarrow$}} & 47.2 & 39.1 \tcbox{\footnotesize{17.2\% $\downarrow$}} \\
         \sciqa{} & 76.8 & 66.1 \tcbox{\footnotesize{13.9\% $\downarrow$}} & 77.5 & 69.6 \tcbox{\footnotesize{10.2\% $\downarrow$}} \\
         \midrule
         \textbf{Average} & 46.6 & 33.6 \tcbox{\footnotesize{27.9\% $\downarrow$}} & 49.2 & 38.3 \tcbox{\footnotesize{22.2\% $\downarrow$}} \\
    \end{tabular}
    \caption{Performance of BART models on \tstp{} vs. \tstnp{} of models trained on data instances containing instruction patterns (\trnp{}).}
\label{app:tab_just_artifacts}
\end{table}
\begin{table}[t]
    \setlength\tabcolsep{4.0pt}
    \setlength{\belowcaptionskip}{-10pt}
    \centering
    \footnotesize
    \begin{tabular}{lcc|cc}
         & \multicolumn{2}{c}{\textbf{Base}} & \multicolumn{2}{c}{\textbf{Large}} \\
         \cmidrule{2-5}
         & \tstp{} & \tstnp{} & \tstp{} & \tstnp{} \\
         \toprule
         \clariq{} & 29.8 & ~~26 ~\tcbox{\footnotesize{12.8\% $\downarrow$}} & 29.4 & 26.3 \tcbox{\footnotesize{10.5\% $\downarrow$}} \\
         \multirc{} & 28.5 & 29.8 ~~\tcbox{\footnotesize{4.6\% $\uparrow$}} & 41.5 & 33.9 \tcbox{\footnotesize{18.3\% $\downarrow$}} \\
         \piqa{} & 22.3 & 20.5 ~~\tcbox{\footnotesize{8.1\% $\downarrow$}} & 23.1 & 21.6 ~~\tcbox{\footnotesize{6.5\% $\downarrow$}} \\
         \quoref{} & 80.5 & 61.2 ~~~\tcbox{\footnotesize{24\% $\downarrow$}} & 87.8 & 73.4 \tcbox{\footnotesize{16.4\% $\downarrow$}} \\
         \ropes{} & ~~44 & 44.1 ~~\tcbox{\footnotesize{0.2\% $\uparrow$}} & 47.9 & 46.5 ~~\tcbox{\footnotesize{2.9\% $\downarrow$}} \\
         \sciqa{} & 76.5 & 70.6 ~~\tcbox{\footnotesize{7.7\% $\downarrow$}} & 52.2 & 50.8 ~~\tcbox{\footnotesize{2.7\% $\downarrow$}} \\
         \midrule
         \textbf{Average} & 46.9 & ~~42 \tcbox{\footnotesize{10.5\% $\downarrow$}} & 47 & 42.1 \tcbox{\footnotesize{10.4\% $\downarrow$}} \\
    \end{tabular}
    \caption{Performance of BART models on \tstp{} vs. \tstnp{} of models trained on \trn{}.}
\label{app:tab_full_resutls}
\end{table}

\section{Dataset Statistics}
\label{app:dataset_stats}

Tab.~\ref{tab:dataset_stats} describes the statistics of train and evaluation sets of datasets used in our experiments. Here, we can observe that each selected dataset differs in terms of number of training samples, \% of instruction patterns, and tasks.

\section{Pattern Extraction Method}
\label{sec:extraction_method}

Here, we describe an example to show how we extract the dominant pattern from the crowdsourcing instructions and subsequently identify the same pattern in the dataset. We try to find recurring word patterns such as \textit{``Are you...''}, \textit{``how many points...''}, \textit{``Was... still...''}, \textit{``since... the...''}. 

For example, \mctaco{} (event duration) has 3 examples in crowdsourcing instructions: (1) how long did Jack play basketball?, (2) how long did he do his homework?, and (3) how long did it take for him to get the Visa? In step (a), we analyze examples manually and find \textit{dominant pattern}. Here, we can see that all examples contain tri-gram pattern, i.e., \textit{``how long did"}. In step (b), we try to generate more possible patterns that are semantically similar to the \textit{dominant pattern} or have a significant word overlap. Here, \textit{``how long did"} can be \textit{``how long was"}, \textit{``how long does"}, etc. (i.e, \texttt{How long AUX}). In step (c), we look for all these possible patterns in datasets using simple word-matching techniques.

\section{Pattern Examples}
\label{app:artifacts_examples}

Tab.~\ref{tab:artifacts_examples} provides dataset, instruction patterns and corresponding examples of data instances that exhibit the instruction patterns.

\begin{table*}[ht]
    \centering
    \setlength\tabcolsep{2.0pt}
    \footnotesize

\resizebox{\linewidth}{!}{

\begin{tabular}{p{1.7cm}p{2.5cm}l}

\textbf{Dataset}        & \textbf{Pattern}                                                               & \textbf{Examples}                                                                              \\ \midrule
\multirow{3}{*}{\clariq}  & \multirow{3}{2.5cm}{\texttt{[Are|Would|Do] you}}                                      &  \texttt{Are you} looking for a specific web site?                                         \\ \cmidrule{3-3} 
                        &                                                                                &  What kind of train \texttt{are you} looking for?                                 \\ \cmidrule{3-3} 
                        &                                                                                & \texttt{Do you} want to watch news videos or read the news? \\ \cmidrule{3-3} 
                        &                                    & \texttt{Would you} like the location of the ritz carlton lake las vegas? \\ \midrule
                        
\multirow{3}{*}{\cosmosqa}  & \multirow{3}{2.5cm}{\texttt{What AUX}}                                      &  \texttt{What may} happen after the young man makes his call?                                         \\ \cmidrule{3-3} 
                        &                                                                                & \texttt{What might} happen if you have him for the whole day?                                  \\ \cmidrule{3-3} 
                        &                                                                                & \texttt{What's} a possible reason the writer doesn't look disabled on the outside?                   \\ \midrule
                        
\multirow{4}{*}{\drop}   & \multirow{4}{2.5cm}{\texttt{How many [field~goals | years | yards | points | touchdowns]}} & \texttt{How many touchdowns} did Jones have?                                                            \\ \cmidrule{3-3} 
                        &                                                                                & \texttt{How many field goals} did Kris Brown kick                                                     \\ \cmidrule{3-3} 
                        &                                                                                & \texttt{How many yards} was the longest touchdown of the game?                                          \\ \cmidrule{3-3} 
                        &                                                                                & After Akers 32-yard field goal, \texttt{how many points} behind was Washington?  \\ \midrule

\multirow{3}{*}{\hotpotqa}  & \multirow{3}{2.5cm}{\texttt{[in|of|from|\_] [Which|What] AUX}}       &  \texttt{Which} franchise \texttt{was} founded in 1978, Chuck E. Cheese's or Jet's Pizza?                                         \\ \cmidrule{3-3} 
                        &                                                                                & Busan, in the area surrounding the mountain of Geumjeongsan, is the second most populated city \texttt{in which} country?                                  \\ \cmidrule{3-3} 
                        &                                                                                & \texttt{What is} the name of the third album from singer Selena Quintanilla-Pérez?                   \\ \midrule

\multirow{5}{*}{\mctaco}  & \multirow{1}{2.5cm}{\texttt{How long AUX}}                                      &  \texttt{How long was} his mother ill?                                         \\ \cmidrule{2-3} 
                        &    \multirow{1}{2.5cm}{\texttt{What AUX}}                                                                            & \texttt{What did} the government decide after the 9/11 attack?                                 \\ \cmidrule{2-3}
                        &     \multirow{1}{2.5cm}{\texttt{How often AUX}}                                                   & \texttt{How often would} one family be able to do something like this?                                  \\ \cmidrule{2-3}
                        &       \multirow{2}{2.5cm}{\texttt{AUX... [still|always] }}        & \texttt{Will} electronic espionage \texttt{always} be happening in the U.S.?                                  \\ \cmidrule{3-3}
                        & & \texttt{Is} she \texttt{still} gone? \\ \cmidrule{2-3}
                        &   \multirow{2}{2.5cm}{\texttt{When did / What time}}          & \texttt{What time} did the planes crash into the World Trade Center?                   \\ \cmidrule{3-3}
                        & & \texttt{When did} Durer die? \\ \midrule

\multirow{3}{*}{\multirc}  & \multirow{3}{2.5cm}{\texttt{What AUX}}                                      &  \texttt{What was} Poe's first published work?                                         \\ \cmidrule{3-3} 
                        &                                                                                & \texttt{What is} the full name of the person described?                                  \\ \cmidrule{3-3} 
                        &                                                                                & \texttt{What} kind of career \texttt{does} Christie Brinkley have?                   \\ \midrule

\multirow{3}{*}{\piqa}  & \multirow{3}{2.5cm}{\texttt{How [do|can]}}                                      &  \texttt{How do} I make orange icing if I have store-bought white frosting?                                          \\ \cmidrule{3-3} 
                        &                                                                                & \texttt{How can} I make popsicles for dogs?                                    \\ \cmidrule{3-3} 
                        &                                                                                & Are you nervous about giving a speech or doing something? \texttt{How can} you calm yourself?                   \\ \midrule
                        
\multirow{4}{*}{\quoref} & \multirow{4}{2.5cm}{\texttt{What AUX~the {[}full | real | first |last{]} name}}                  & \texttt{What is the first name} of the person who purchases a revolver?                                 \\ \cmidrule{3-3} 
                        &                                                                                & \texttt{What is the full name} of the person who is calmly asked to leave?                              \\ \cmidrule{3-3} 
                        &                                                                                & \texttt{What was the name} of the house where Appleton Water Tower was built?                           \\ \cmidrule{3-3} 
                        &                                                                                & \texttt{What is the last name} of the person who convinces the girls to help him look for the treasure? \\ \midrule

\multirow{3}{*}{\ropes}  & \multirow{3}{2.5cm}{\texttt{Which AUX}}                                      &  \texttt{Which} area \texttt{would} be less likely to experience a drought and have better chance at a new growth?                                          \\ \cmidrule{3-3} 
                        &                                                                                & \texttt{Which} hair spray brand \texttt{should} Greg buy to be environmentally friendly?                                    \\ \cmidrule{3-3} 
                        &                                                                                & \texttt{Which} markalong \texttt{was} produced asexually?                   \\ \midrule
                        
\multirow{3}{*}{\sciqa}  & \multirow{3}{2.5cm}{\texttt{What AUX}}                                                      & \texttt{What are} by far the most common type of invertebrate?                                          \\ \cmidrule{3-3} 
                        &                                                                                & \texttt{What do} waves deposit to form sandbars and barrier islands?                                    \\ \cmidrule{3-3} 
                        &                                                                                & \texttt{What is} the term for the total kinetic energy of moving particles of matter?                   \\ \midrule

\multirow{3}{1.6cm}{\textsc{Wino- grande}}  & \multirow{3}{2.5cm}{\texttt{[because | so | while | since | but] ... the}}                &    The dog didn't like its collar but was okay with its leash \texttt{because the} \_ was loose on it.                                      \\ \cmidrule{3-3} 
                        &                                                                                & Hunter took Benjamin's clothes to the laundromat, \texttt{since} \_ had \texttt{the} day off that day. \\ \cmidrule{3-3}
                        &              & James sang his song at the top of his voice so as to be heard over the noise \texttt{but the} \_ is louder.                                       \\ \bottomrule
\end{tabular}
}
\caption{Examples of data instances from original dataset that contain instruction patterns. $\texttt{AUX} \in \{$am, is, are, was, were, has, have, had, do, does, did, will, would, can, could, may, might, shall, should, must$\}$. \_ : <blank>.}
\label{tab:artifacts_examples}
\end{table*}


\section{The Effect of Instruction Examples on Pattern Frequency in Collected Data}
\label{app:survey}

To study the effect of bias in instruction examples on collected data, 
we asked NLP graduate students to write five questions for each of (1) temporal reasoning (event duration) and (2) coreference resolution, based on the crowdsourcing instructions of \mctaco{} and \quoref{}, respectively. For each task, we conduct two surveys, where the instructions include and do not include any examples.

We collected responses from 10 participants.
The dominant patterns of \mctaco{} (`how long') and \quoref{} (`What is the name') only contribute to 38\% and 8\% of our collected data where examples are not given, in contrast to 68\% ($\uparrow$79\%) and 32\% ($\uparrow$300\%) in collected data where examples are given. This indicates that crowdsourcing examples bias crowdworkers to follow certain patterns, whereas showing no examples increases the creativity of crowdworkers.

In addition, our collected responses where examples are not given contain 10 and 9 unique patterns for \mctaco{} (event duration) and \quoref{} respectively, in contrast to only 4 and 5 unique patterns in collected data where examples are given. Our finding shows that there is substantial linguistic diversity associated with the NLP tasks, unlike the patterns covered in instruction examples that get propagated to corresponding datasets. 


The task instructions and collected annotations are available at \url{https://github.com/Mihir3009/instruction-bias/blob/main/SURVEY.md}.




\section{Additional Results}
\label{app:additional_results}

Tab.~\ref{app:tab_just_artifacts} and Tab.~\ref{app:tab_full_resutls} show the performance of BART on \tstp{} and \tstnp{} when training only on examples with instruction patterns and the full training set, respectively. From Tab.~\ref{app:tab_just_artifacts}, there are large performance gaps reaching $45.9\%$ in \multirc{} and $>20\%$ in both base and large models for \quoref{}, and \piqa{}. Overall, the average performance across all datasets is $27.9\%$ and $22.2\%$ higher on \tstp{} for the base and large models, respectively. This indicates that both base and large models often fail to generalize beyond instruction patterns.

From Tab.~\ref{app:tab_full_resutls}, we see that the average performance across all datasets is higher on examples that exhibit instruction patterns by $\sim10.5\%$ for both base and large models. From the results, we can conclude that the model's performance is overestimated by instruction bias.

\end{document}